\documentclass[sigconf]{acmart}

\copyrightyear{2025}

\acmYear{2025}

\setcopyright{rightsretained}

\acmConference[KDD '25] {Proceedings of the 1st Workshop on "AI for Supply Chain: Today and Future" @ 31st ACM SIGKDD Conference on Knowledge Discovery and Data Mining V.2}{August 3, 2025}{Toronto, ON, Canada.}

\acmBooktitle{Proceedings of the 1st Workshop on "AI for Supply Chain: Today and Future" @ 31st ACM SIGKDD Conference on Knowledge Discovery and Data Mining V.2 (KDD '25), August 3, 2025, Toronto, ON, Canada}

\acmISBN{979-8-4007-1454-2/25/08}

\acmDOI{10.1145/XXXXXX.XXXXXX}

\begin{document}

\title{Intelligent Routing for Sparse Demand Forecasting: A Comparative Evaluation of Selection Strategies}

\author{Qiwen Zhang}
\affiliation{%
  \institution{Independent Researcher}
  \city{Atlanta}
  \state{Georgia}
  \country{USA}}
\email{owenzhang1999@gmail.com}

\renewcommand{\shortauthors}{Zhang}

\begin{abstract}
Sparse and intermittent demand forecasting in supply chains presents a critical challenge, as frequent zero-demand periods hinder traditional model accuracy and impact inventory management. We propose and evaluate a Model-Router framework that dynamically selects the most suitable forecasting model—spanning classical, ML, and DL methods—for each product based on its unique demand pattern. By comparing rule-based, LightGBM, and InceptionTime routers, our approach learns to assign appropriate forecasting strategies, effectively differentiating between smooth, lumpy, or intermittent demand regimes to optimize predictions. Experiments on the large-scale Favorita dataset show our deep learning (InceptionTime) router improves forecasting accuracy by up to 11.8\% (NWRMSLE) over strong, single-model benchmarks with 4.67x faster inference time. Ultimately, these gains in forecasting precision will drive substantial reductions in both stockouts and wasteful excess inventory, underscoring the critical role of intelligent, adaptive AI in optimizing contemporary supply chain operations.
\end{abstract}



\maketitle

\section{Introduction}
Accurate demand forecasting is pivotal in supply chain management, directly influencing inventory control, procurement strategies, and overall operational efficiency. However, sparse and intermittent demand patterns, characterized by irregular occurrences and prolonged periods of zero demand, present significant forecasting challenges. Traditional forecasting methods such as exponential smoothing and simple regression models typically assume continuous or regular demand patterns, which limits their effectiveness in sparse scenarios\cite{croston1972forecasting}.

Addressing this limitation, recent advancements have introduced specialized approaches such as Croston's method \cite{croston1972forecasting} and various machine learning (ML) and deep learning (DL) algorithms. Nevertheless, due to the diversity in sparsity patterns across different products and stores, no single forecasting model consistently performs best across all scenarios \cite{m3_competition}. Therefore, a promising direction is to dynamically select the optimal forecasting method tailored to each time series.

In this paper, we address the challenge of selecting the optimal forecasting model by developing and comparing three distinct meta-model routing approaches. These routers are designed to intelligently select the most suitable forecasting method for each individual demand series from a diverse model bank. This model bank was strategically assembled to include models with complementary strengths, spanning classical methods tailored for specific patterns (e.g., Croston\cite{croston1972forecasting}, ETS\cite{HYNDMAN2002439}), a robust ML model (LightGBM\cite{ke2017lightgbm}), and state-of-the-art DL approaches capable of learning complex dependencies (DeepAR\cite{salinas2020deepar}, PatchTST\cite{nie2023patchtst}). The necessity of routing among these models stems from the well-documented observation that no single forecasting method consistently outperforms others across the wide variety of demand patterns found in real-world retail data \cite{m3_competition}. Our investigated routing mechanisms include: (1) a rule-based system leveraging specific demand characteristics (variability and autocorrelation), (2) a machine learning router (LightGBM) trained on a comprehensive set of extracted time series features, and (3) a deep learning router (InceptionTime\cite{Ismail_Fawaz_2020}) which operates directly on the raw time series data. We extensively evaluate these routing strategies using the Favorita\cite{Kaggle} dataset, demonstrating their effectiveness and potential for significant accuracy improvements over conventional single-model forecasting methods.

The main contributions of our study include:
\begin{itemize}
    \item A comprehensive feature extraction strategy specifically designed to capture sparsity and intermittency in demand data.
    \item Implementation of a novel model-routing framework incorporating rule-based, ML-based (LightGBM), and DL-based (InceptionTime) selection mechanisms.
    \item Empirical validation on a large-scale dataset, highlighting significant forecasting accuracy improvements.
\end{itemize}

\section{Related Work}
Forecasting intermittent and sparse demand has been extensively studied in supply chain management literature. Croston's method \cite{croston1972forecasting} remains a classical and widely-adopted approach specifically designed to handle intermittent demand by separately forecasting demand size and intervals between demands.

Recent advances in machine learning and deep learning have further expanded the toolkit for sparse demand forecasting. Models such as LightGBM \cite{ke2017lightgbm} have demonstrated strong performance due to their efficiency and robustness in handling heterogeneous data. Deep learning models, including recurrent neural networks (RNNs\cite{RNN}), attention-based models such as DeepAR \cite{salinas2020deepar} and PatchTST \cite{nie2023patchtst}, leverage temporal dependencies and complex patterns within the data to enhance forecasting accuracy. Furthermore, architectures like InceptionTime \cite{Ismail_Fawaz_2020} have shown exceptional performance in time series classification tasks by effectively capturing features at different temporal scales, a capability that can be adapted for model selection based on time series characteristics.

However, given the heterogeneous nature of sparsity across different products and stores, reliance on a single forecasting approach often proves insufficient \cite{m3_competition}. This has motivated research into meta-learning and model-routing frameworks, which dynamically select optimal forecasting models based on specific data characteristics. Recent studies \cite{talagala2018meta, montero2020fforma} have employed meta-learning techniques for time series forecasting, highlighting improvements by selecting forecasting methods tailored to individual time series.

Despite these advancements, the comparative performance of distinct rule-based, machine learning-based, and deep learning-based routing strategies, particularly when applied to the challenge of sparse demand forecasting, has not been comprehensively explored. Our study addresses this by systematically evaluating these three individual model-routing approaches, offering insights into their respective strengths and weaknesses in this domain.

\section{Methodology}
\begin{figure}[htbp]
    \centering
    \includegraphics[width=0.8\linewidth]{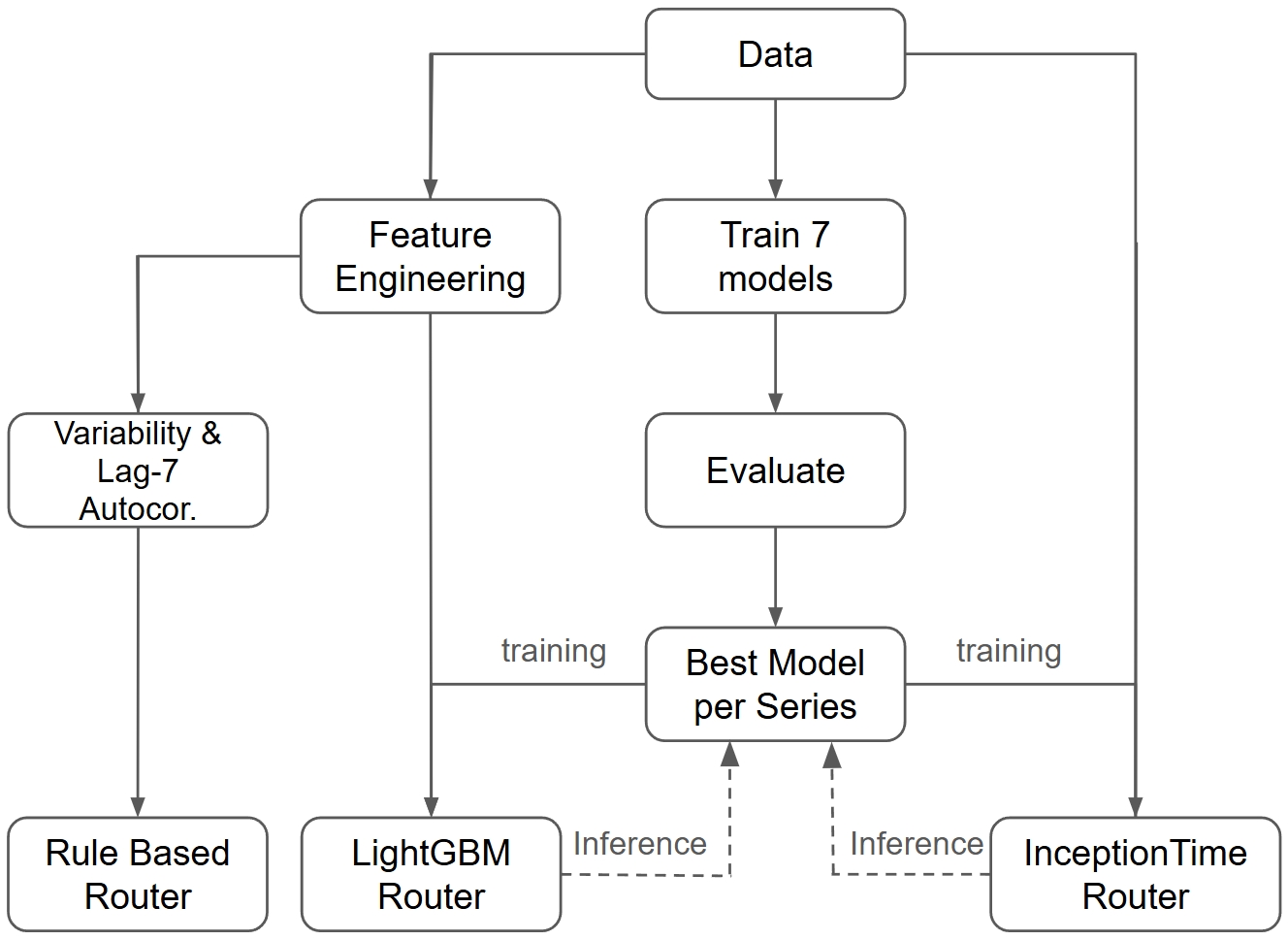}
    \caption{Model routing overview. Data flows into (1) model training and evaluation for each time series to generate the \textit{Best Model per Series} target, and (2) feature engineering as the input for the \textbf{LightGBM} and \textbf{Rule-Based} routers. The \textbf{InceptionTime} router uses raw time series as input; both it and LightGBM Router predict the \textit{Best Model per Series} as their target.}
    \label{fig:overall_process}
\end{figure}

The overall methodology, from data preparation to the evaluation of different model routing strategies, is depicted in Figure~\ref{fig:overall_process}.

\subsection{Problem Definition}
We define the sparse demand forecasting problem as follows: given historical daily sales data for product-store combinations, our goal is to forecast future sales accurately, despite intermittent zero-demand periods and high variability. Formally, for each time series \( \{y_t\}_{t=1}^T \), we aim to predict future values \( \{y_{T+1}, ..., y_{T+h}\} \), where \(h\) denotes the forecast horizon.

Given the diverse nature of these series and the observation that no single forecasting model excels across all of them, we introduce a secondary, meta-problem: \textbf{the model selection task}. For each series \(i\), our objective is to identify the optimal forecasting model, \( M_i^* \), from a pre-defined bank of candidate models \( \mathcal{M} = \{M_1, ..., M_k\} \). The goal is to select \( M_i^* \) such that it minimizes a given forecasting error metric (in our case, NWRMSLE) for the horizon \(h\). Our proposed model routers aim to learn a function \( f \) such that \( M_i^* \approx f(\text{characteristics}_i) \) or \( M_i^* \approx f(\{y_t\}_{t=1}^T) \).

\subsection{Data Preparation and Feature Engineering}\label{sec:data_prep_features}
We utilize the Favorita dataset\cite{Kaggle}, sourced from a Kaggle competition, consisting of daily sales data at the most granular level (product-store combinations). We define a unique identifier \texttt{series\_id} combining store and item numbers. We randomly sampled 19,646 time series from the Favorita dataset for model training and meta-model router development. Additionally, we sampled 5,000 unique time series as an independent holdout set for offline evaluation, ensuring these series were not observed during the training phase. This approach ensures unbiased evaluation and generalizability of our model-routing strategy.

For each time series, we extract comprehensive features using the \texttt{tsfresh} library and additional custom features capturing sparsity characteristics.

The features extracted from \texttt{tsfresh} include basic statistics (mean, variance), seasonality and trend features (FFT coefficients), autocorrelation, entropy measures, and trend indicators. Additionally, we calculate custom sparsity-focused features:
\begin{itemize}
    \item Demand frequency (percentage of non-zero demand)
    \item Intermittency (mean interval between non-zero demands)
    \item Variability (coefficient of variation)
    \item Mean demand
\end{itemize}

This comprehensive feature extraction yields a total of 34 features for each time series.

\subsection{Data Overview}
\begin{figure}[htbp]
    \centering
    \includegraphics[width=0.8\linewidth]{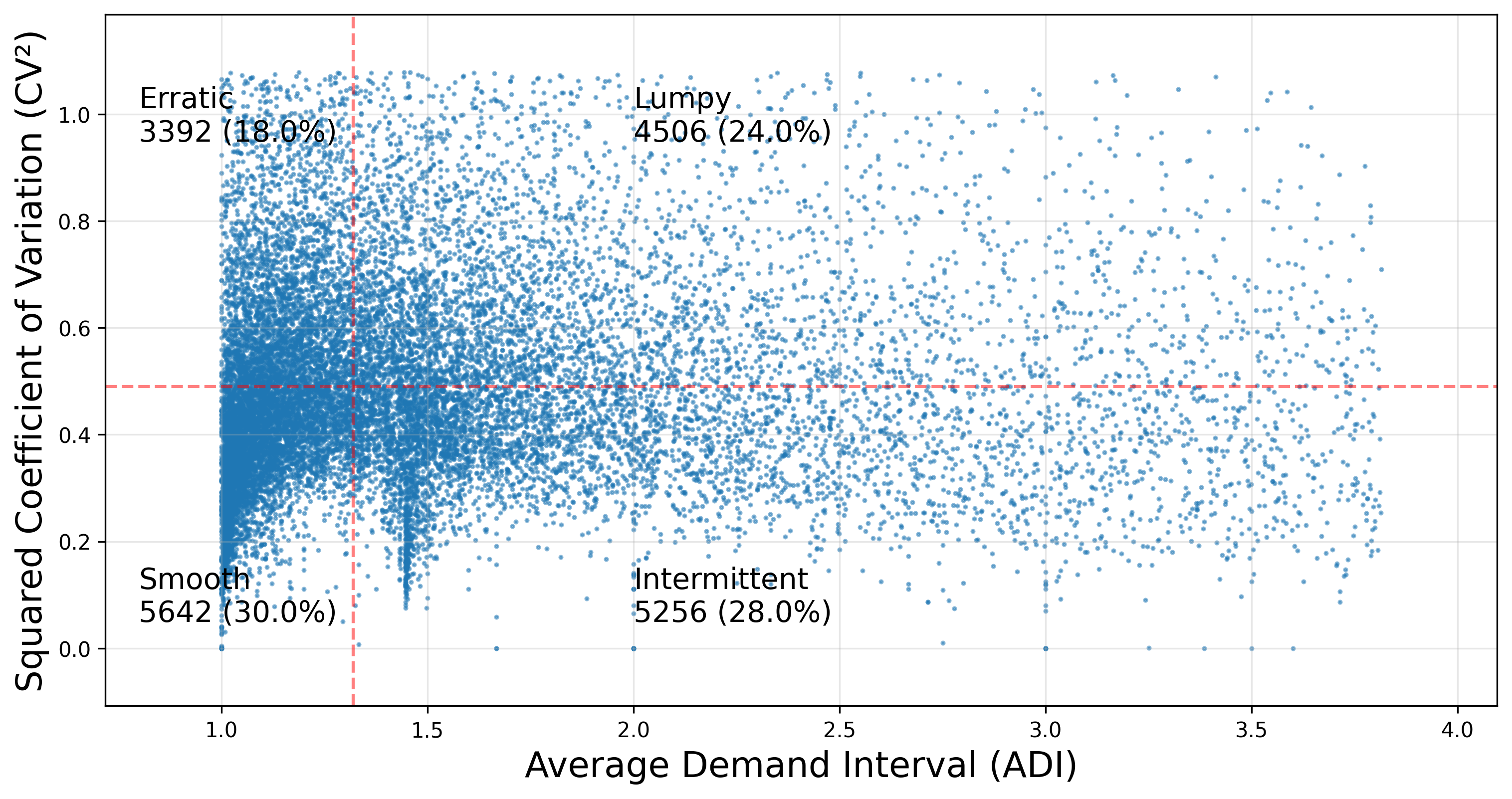}
    \caption{ADI vs. CV\textsuperscript{2} classification of demand patterns.}
    \label{fig:adi_cv_placeholder}
\end{figure}

A thorough exploratory data analysis (EDA) was conducted on the prepared dataset, comprising 24,646 time series (19,646 for training and 5,000 for holdout evaluation), to understand its inherent characteristics, particularly concerning demand sparsity and feature interactions. Key findings from this EDA informed subsequent modeling choices.
\begin{itemize}
    \item \textbf{Feature Correlation Analysis:} To assess the relationships and redundancy among the 34 engineered time series features (Section~\ref{sec:data_prep_features}), a correlation heatmap was generated (see Figure~\ref{fig:heatmap_placeholder}). The heatmap visually represents the Pearson correlation coefficients between all pairs of features. The analysis indicated that while some features exhibited expected correlations (e.g., different lags of autocorrelation), the majority of features displayed low to moderate correlations with each other. This general lack of strong multicollinearity validated the feature set's diversity, suggesting that the features capture distinct aspects of the time series dynamics, which is beneficial for training robust machine learning models for the routing task.
\begin{figure}[htbp]
    \centering
    \includegraphics[width=0.8\linewidth]{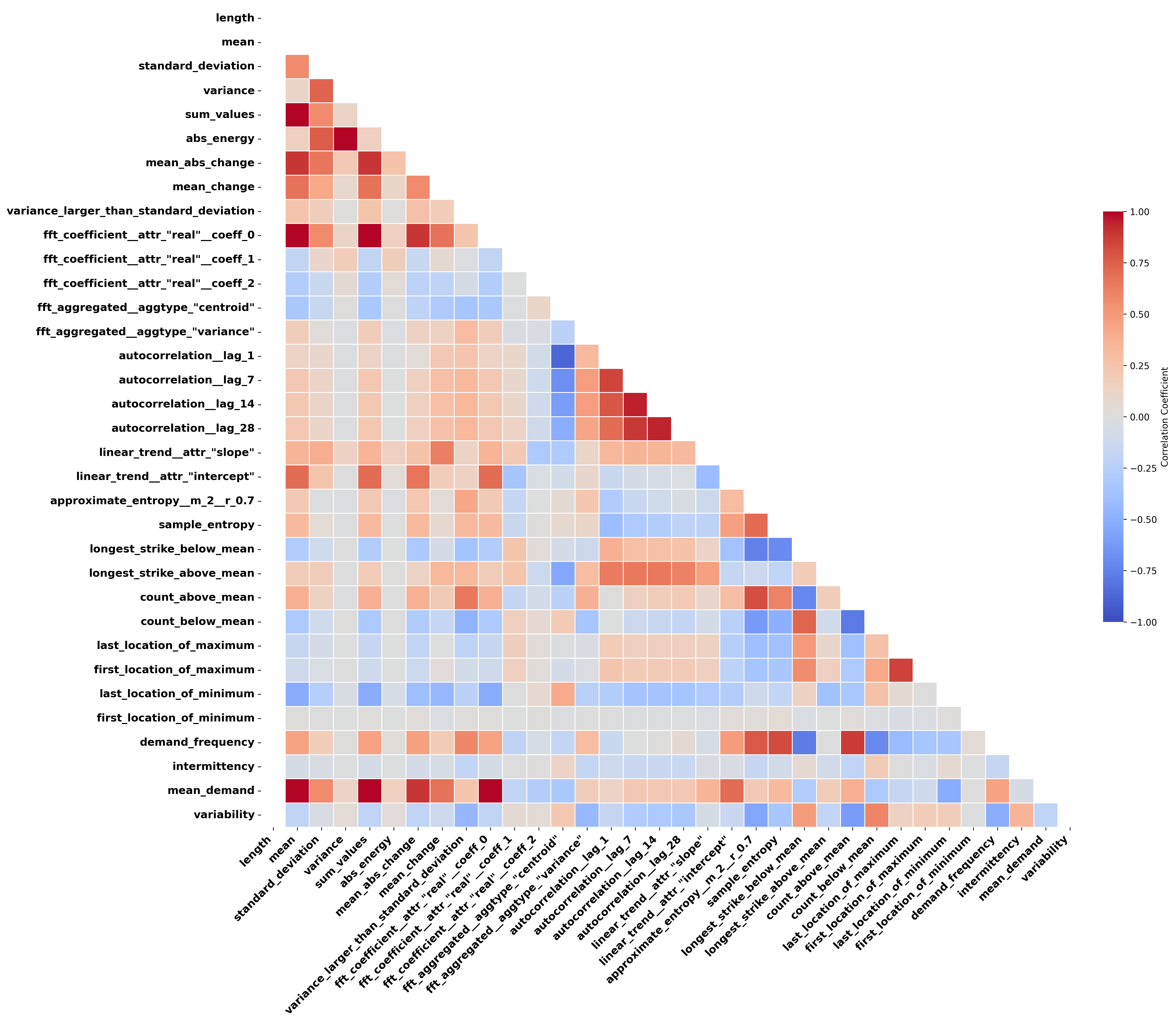}
    \caption{Correlation heatmap of extracted features.}
    \label{fig:heatmap_placeholder}
\end{figure}

    \item \textbf{Distribution of Zero-Sales Percentages:} The prevalence of sparsity was quantified by analyzing the distribution of zero-sales percentages for each individual time series. A histogram of these percentages (see Figure~\ref{fig:zero_sales}) revealed a significant level of sparsity across the dataset. The mean percentage of zero sales per time series was found to be 57.2\%, with a median of 62.2\%. Such high figures underscore the intermittent nature of the demand, posing a substantial challenge for traditional forecasting models that assume more continuous data and highlighting the necessity for models adept at handling periods of no demand.

\begin{figure}[htbp]
    \centering
    \includegraphics[width=0.8\linewidth]{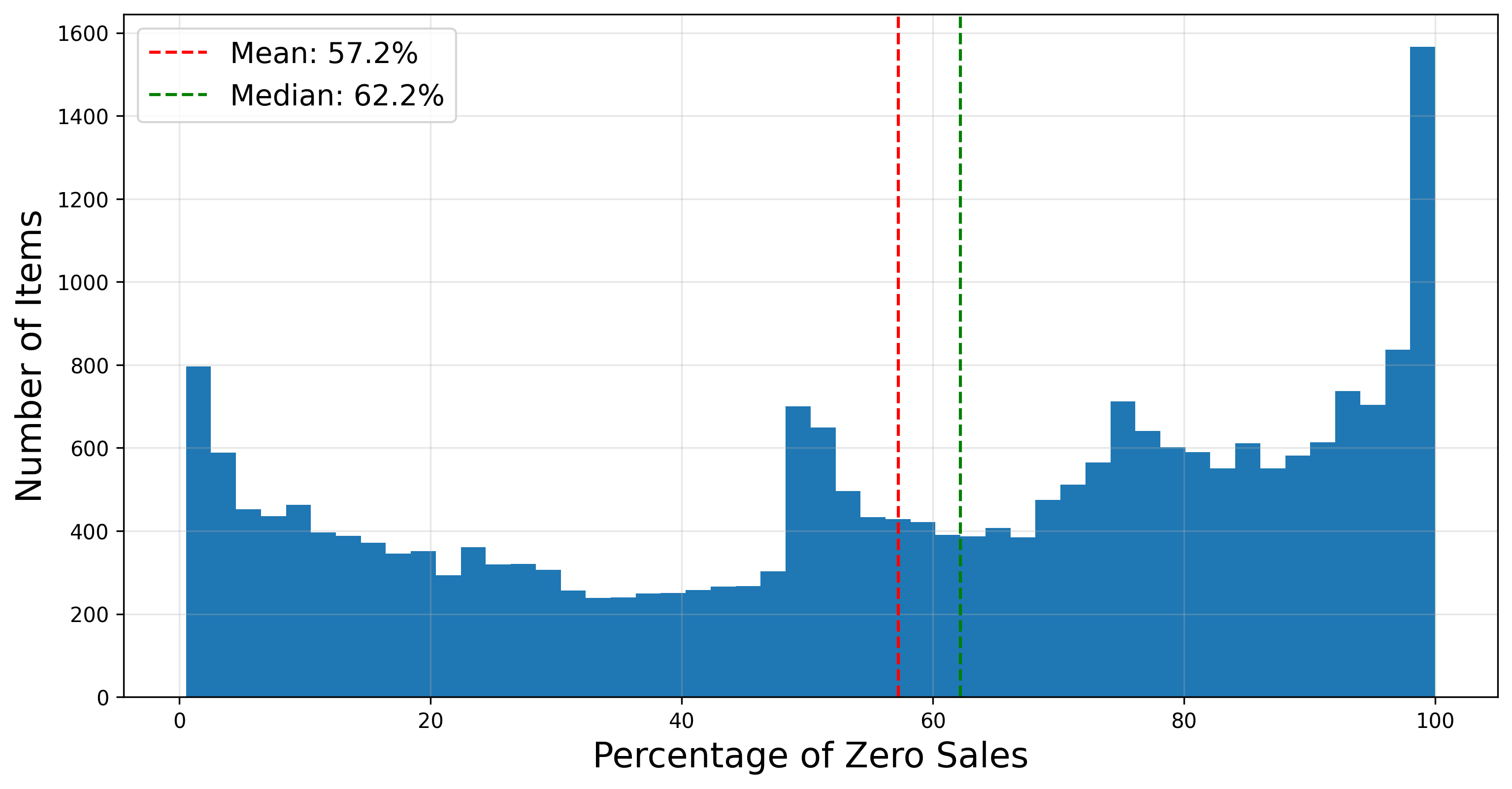}
    \caption{Distribution of Zero Sales Percentage}
    \label{fig:zero_sales}
\end{figure}

    \item \textbf{Demand Pattern Classification (ADI vs. CV\textsuperscript{2}):} To further characterize the demand patterns, a scatter plot of Average Demand Interval (ADI) versus the squared Coefficient of Variation (CV\textsuperscript{2}) was utilized, following established inventory management literature for classifying demand \cite{syntetos2005accuracy}. Outliers were removed using the Interquartile Range (IQR) method prior to plotting to ensure a clearer representation of the dominant patterns. The analysis (see Figure~\ref{fig:adi_cv_placeholder}) categorized the time series into four distinct demand patterns: smooth (30\% of series), erratic (18\%), lumpy (24\%), and intermittent (28\%). The significant presence of erratic, lumpy, and intermittent patterns (collectively 70\% of the data) further confirmed the dataset's sparse and irregular nature, reinforcing the need for a model selection strategy that can adapt to such diverse behaviors.
\end{itemize}

\subsection{Model Bank Formulation and Evaluation}

\subsubsection{Rationale for Model Selection}
A critical component of our framework is the model bank, which serves as a pool of candidate experts for the router. The selection of models was guided by the principle of **complementarity**, aiming to include a diverse set of methods, each with distinct theoretical underpinnings and strengths suited to different types of time series patterns observed in sparse demand data. This diversity is essential for the routing concept to be effective, as it ensures that for any given time series, a potentially suitable 'expert' exists within the bank. Our goal was not to create an exhaustive list of all possible forecasting models, but rather to construct a representative and varied pool to rigorously test the efficacy of different routing strategies.

The model bank comprises seven forecasting methods, categorized as follows:
\begin{itemize}
    \item \textbf{Classical Statistical Models:} This group forms a foundation of well-understood and interpretable methods.
    \begin{itemize}
        \item \textbf{Exponential Smoothing (ETS):} Chosen for its robustness and effectiveness in capturing underlying trend and seasonality, making it a strong candidate for more regular, 'smooth' demand patterns \cite{hyndman2018forecasting}.
        \item \textbf{Croston's Method:} Included specifically for its theoretical design to handle intermittent demand. It separately models the time between demands and the demand size, making it a specialized expert for 'intermittent' and 'lumpy' series \cite{croston1972forecasting}.
        \item \textbf{Naïve and Moving Average:} These serve as essential, computationally inexpensive baselines. Their inclusion provides a "simple but robust" option, which can often outperform more complex models on highly erratic or unpredictable series.
    \end{itemize}
    \item \textbf{Machine Learning Model:}
    \begin{itemize}
        \item \textbf{LightGBM:} Represents the class of gradient boosting machines, which have proven highly effective in forecasting competitions \cite{ke2017lightgbm}. It was chosen for its ability to model complex, non-linear relationships between a rich set of engineered time series features (as described in Section~\ref{sec:data_prep_features}) and the forecast target. It acts as a feature-based expert in the bank.
    \end{itemize}
    \item \textbf{Deep Learning Models:} This category includes modern, state-of-the-art architectures capable of learning representations directly from raw time series.
    \begin{itemize}
        \item \textbf{DeepAR:} A probabilistic forecasting model based on recurrent neural networks (RNNs). It was selected for its ability to learn global models from large panels of related time series and handle temporal dependencies \cite{salinas2020deepar}.
        \item \textbf{PatchTST:} A Transformer-based model that has recently demonstrated state-of-the-art performance on long-sequence forecasting tasks \cite{nie2023patchtst}. Its inclusion tests whether cutting-edge attention mechanisms, which analyze data in 'patches', provide a distinct advantage for certain series within our dataset.
    \end{itemize}
\end{itemize}

While this selection represents a strong and diverse pool, we acknowledge that the efficacy of the routing system is inherently tied to the strength of its constituent models. For instance, the deep learning models (DeepAR and PatchTST) were trained with a limited number of iterations (100) due to computational constraints. While this may underestimate their full potential, it creates a realistic testbed for the router, whose task is to select the best model \textbf{given the available trained instances}. The primary objective of this study is to demonstrate the value of the routing mechanism itself, and this curated bank provides the necessary diversity to validate that concept.

\subsubsection{Best Model Identification for Router Training}
With the model bank established, the next step is to generate the ground-truth labels required for training our supervised learning routers (LightGBM and InceptionTime). The primary objective of the model bank at this stage is not to forecast directly, but to serve as a pool of experts from which we can determine the optimal model for each time series\textit{ a priori} within our training set (19,646 series).

To achieve this, each of the seven models was individually trained and evaluated on every time series. This evaluation was conducted for two separate forecasting horizons: 14 days and 30 days.

Forecast accuracy was evaluated using the Normalized Weighted Root Mean Squared Logarithmic Error (NWRMSLE). We specifically selected this metric because it was the official evaluation criterion for the original Kaggle ``Corporación Favorita Grocery Sales Forecasting'' competition, ensuring our performance assessment aligns with the dataset's established benchmark. The NWRMSLE is defined as:
\begin{equation}
\text{NWRMSLE} = \sqrt{\frac{\sum_{i=1}^n w_i\left(\ln(\hat{y}_i+1)-\ln(y_i+1)\right)^2}{\sum_{i=1}^n w_i}}
\label{eq:nwrmsle}
\end{equation}
where $n$ is the number of forecast points, $\hat{y}_i$ is the predicted sales, $y_i$ is the actual sales, and $w_i$ is the weight. Following the competition's rules, $w_i$ is set to 1.25 for perishable items and 1.00 for all others, giving higher importance to accurately forecasting perishable goods.

For each time series and each forecast horizon, the model achieving the lowest NWRMSLE was identified and recorded. This created our "best model" label, a categorical feature indicating the most suitable forecasting method (out of the seven) for a specific series and horizon. This label serves as the ground truth and the target variable for training our classification-based model routers (LightGBM and InceptionTime), as detailed in the following section.

\subsection{Model Routing Approaches}
To dynamically select the most appropriate forecasting model from the bank for each time series, we developed and evaluated three distinct routing strategies. These strategies range from a simple heuristic approach to more complex machine learning and deep learning models. For the LightGBM and InceptionTime routers, separate models were trained for the 14-day and 30-day forecast horizons.

\begin{itemize}
    \item \textbf{Rule-Based Router:}
    This approach employs predefined rules based on two key time series characteristics: variability (coefficient of variation, \texttt{variability}) and weekly seasonality (autocorrelation at lag 7, \texttt{value\_autocorrelation\_lag\_7}). The rules are:
    \begin{itemize}
        \item If \texttt{variability < 0.8} (relatively stable demand):
        \begin{itemize}
            \item If \texttt{value\_autocorrelation\_lag\_7 > 0.3} (strong weekly seasonality): Route to \textbf{ETS}. ETS models are well-suited for series with clear trend and seasonal components \cite{hyndman2018forecasting}.
            \item Else (weak/no weekly seasonality): Route to \textbf{Moving Average}. For stable series without strong seasonality, a simple Moving Average can provide robust forecasts by smoothing out noise \cite{hyndman2018forecasting}.
        \end{itemize}
        \item Else (\texttt{variability >= 0.8}, more variable demand):
        \begin{itemize}
            \item If \texttt{value\_autocorrelation\_lag\_7 > 0.4} (strong temporal dependence): Route to \textbf{PatchTST}. PatchTST, as a Transformer-based model, is designed to capture complex temporal dependencies and can be effective for variable series that still exhibit underlying patterns.
            \item Else (high variability, no clear structure): Route to \textbf{Moving Average}. This serves as a conservative baseline for highly unpredictable series, as it was observed to be effective in preliminary analyses.
        \end{itemize}
    \end{itemize}
    The selected models (ETS, Moving Average, PatchTST) align with these heuristically defined segments of time series characteristics.

    \item \textbf{LightGBM Router:}
    A machine learning router using LightGBM, known for its efficiency and performance on tabular data. It uses the 34 engineered time series features (see Section~\ref{sec:data_prep_features}) as input. The target is the "best model" label from the model bank evaluation. LightGBM was chosen for its ability to handle heterogeneous features, model non-linearities, and address class imbalance using the \texttt{class\_weight} parameter.

    \item \textbf{InceptionTime Router:}
    A deep learning router using InceptionTime \cite{Ismail_Fawaz_2020}, a strong performer in time series classification. It processes raw time series data directly, enabling end-to-end learning for the routing task. InceptionTime predicts the "best model" label by learning discriminative features across multiple temporal scales.
\end{itemize}

\section{Experimental Setup}

\subsection{Model Bank Training Configurations}
\label{sec:model_bank_training}

The seven forecasting models comprising our model bank were trained to generate the 'best model' labels for each time series and horizon. The training configurations and sources were as follows:

\begin{itemize}
    \item \textbf{Classical Models:}
        \begin{itemize}
            \item Exponential Smoothing (ETS) was implemented using the \texttt{statsmodels} Python package \cite{seabold2010statsmodels}. 
            \item Croston's method was implemented using the \texttt{croston} Python package \cite{croston2021}. 
            \item Naïve (last value) and Moving Average methods were custom-implemented for this study.
        \end{itemize}
    \item \textbf{LightGBM (as forecaster):} When used as a direct forecasting model within the bank, LightGBM was implemented using the official \texttt{lightgbm} Python package \cite{ke2017lightgbm} and trained using its default hyperparameters for a regression task.
    \item \textbf{Deep Learning Models:} Both DeepAR and PatchTST were implemented using the \texttt{neuralforecast} Python package \cite{olivares2022library_neuralforecast}. 
    They were trained using their respective standard architectures but were constrained to only 100 training iterations. This limitation was necessary due to computational resource constraints, balancing model potential with practical feasibility across thousands of series.
\end{itemize}

These configurations ensured that we could establish a baseline performance for each model type across all series, which is crucial for generating meaningful target labels for the subsequent router training.

\subsection{Model Router Training Configurations}
\label{sec:model_router_training}

The meta-models, or routers, were trained as classifiers to predict the 'best model' label identified in the previous step. Two primary routers were developed:

\begin{itemize}
    \item \textbf{LightGBM Router:} This model was trained as a multi-class classifier using the \texttt{lightgbm} package \cite{ke2017lightgbm}. It utilized the 34 engineered time series features (see Section~\ref{sec:data_prep_features}) as input. While most hyperparameters were kept at their default settings, the \texttt{class\_weight='balanced'} parameter was crucial. This setting was enabled to counteract the inherent class imbalance, ensuring that the router did not become biased towards predicting only the most frequently occurring 'best models'.
    \item \textbf{InceptionTime Router:} This deep learning router was also trained as a multi-class classifier, taking the raw time series as input. It employed a standard training regimen suitable for deep time series classification, including an appropriate optimizer (e.g., Adam) and loss function (e.g., categorical cross-entropy). Similar to the LightGBM router, strategies to handle potential class imbalance were considered during its training process.
\end{itemize}

Separate instances of both the LightGBM and InceptionTime routers were trained for the 14-day and 30-day forecasting horizons, as the 'best model' labels can differ depending on the forecast length.


\subsection{Benchmarking Model Selection}

To effectively evaluate the performance gains achieved by our proposed model routing strategies, it is crucial to establish strong baselines. We achieve this through \textbf{single-model benchmarking}. This process involves selecting a single, representative forecasting model and applying it uniformly across \textbf{all} 5,000 time series in our holdout set. We then calculate the overall NWRMSLE achieved by this single model.

This approach provides a clear answer to the question: ``How well would we perform if we simply chose one 'best' or 'representative' model and used it everywhere, without any dynamic selection?'' By comparing the NWRMSLE achieved by our model routers against these single-model benchmarks, we can directly quantify the value added by our intelligent, series-specific model selection process.

For this benchmarking, we selected three models representing different forecasting paradigms present in our model bank:

\begin{itemize}
    \item \textbf{Croston:} A classical method specifically designed for sparse and intermittent demand, known for its specialized handling of zero-demand intervals \cite{croston1972forecasting}.
    \item \textbf{LightGBM:} A powerful machine learning model, representing gradient boosting approaches known for their efficiency and predictive strength on structured data \cite{ke2017lightgbm}.
    \item \textbf{PatchTST:} A state-of-the-art deep learning model, representing Transformer-based architectures that can effectively capture complex temporal dependencies \cite{nie2023patchtst}.
\end{itemize}

These models were chosen to provide a comprehensive and challenging set of benchmarks, covering traditional, ML-based, and DL-based forecasting methods, against which our routing strategies can be rigorously evaluated.

\subsection{Forecasting Performance Evaluation}
\label{sec:forecasting_evaluation}

The forecasting efficacy of our model routers was evaluated on an independent holdout set of 5,000 time series. Performance was measured using the Normalized Weighted Root Mean Squared Logarithmic Error (NWRMSLE), as defined in Equation~\ref{eq:nwrmsle}. We compared the NWRMSLE achieved by our Rule-Based, LightGBM, and InceptionTime routers against three single-model benchmarks: Croston, LightGBM (as a forecaster), and PatchTST, each applied uniformly to all series. This evaluation was conducted for both 14-day and 30-day forecast horizons, with results analyzed overall and segmented by demand patterns (smooth, erratic, intermittent, lumpy).

\subsection{Model Router Classification Performance}
\label{sec:router_accuracy_concise}

To assess how well the LightGBM and InceptionTime routers identify the optimal forecasting model for each series, we evaluated their classification performance on a validation set. The metrics used were overall accuracy, macro-average F1 score, and weighted-average F1 score. The F1 score, being the harmonic mean of precision and recall, offers a balanced measure, while its macro and weighted variants provide insights into performance across potentially imbalanced classes (i.e., different 'best models'). These metrics were reported for routers trained for both 14-day and 30-day horizons.

\section{Results}

\subsection{Forecasting Accuracy Results}
\label{sec:forecasting_performance}
The forecasting performance of the single-model benchmarks and our proposed model routers on the 5,000-series holdout set is summarized in Table~\ref{tab:forecasting_performance}. The results clearly demonstrate a substantial advantage in employing adaptive model selection via routers, as all three router configurations consistently outperformed the all of the single-model approachs across both 14-day and 30-day horizons.

Overall, the \textbf{InceptionTime router} achieved the lowest NWRMSLE, with scores of \textbf{2.051} for the 14-day horizon and \textbf{3.134} for the 30-day horizon. This represents an improvement of approximately 11.7\% and 11.8\%, respectively, compared to the strongest single-model benchmark, PatchTST (NWRMSLE of 2.323 and 3.555). The LightGBM router also showed strong performance with overall NWRMSLEs of 2.168 (14-day) and 3.282 (30-day), followed by the Rule-based router (2.209 and 3.382).

The superior performance of the InceptionTime router, which operates directly on raw time series data, underscores the potential of end-to-end deep learning models to automatically extract salient features for complex model selection tasks in forecasting. This router demonstrated robustness by achieving the best NWRMSLE across all individual demand patterns (Smooth, Intermittent, Lumpy, and Erratic) for both forecast horizons. For instance, in forecasting \texttt{Intermittent} demand, a key challenge in sparse scenarios, the InceptionTime router achieved NWRMSLEs of 1.748 (14-day) and 2.633 (30-day), outperforming all other methods. Similarly, for \texttt{Smooth} demand, it achieved scores of 2.053 and 3.085. These findings align with a growing body of research highlighting the efficacy of deep learning techniques for recognizing and modeling complex, non-linear patterns in intermittent and sparse demand data \cite{dl_review_2025, dl_review_2022}. 

While single models like PatchTST performed competitively, particularly for `Intermittent` and `Lumpy` patterns compared to other single models, they were consistently surpassed by the dynamic selection capabilities of the routers. This highlights the inherent limitation of a one-size-fits-all forecasting strategy when dealing with heterogeneous time series datasets typical in supply chain management.

\begin{table}[htbp]
\centering
\caption{Forecasting Performance (NWRMSLE) by Demand Pattern and Overall. Best performance in each category is highlighted in bold.}
\resizebox{\linewidth}{!}{%
\begin{tabular}{llcc}
\toprule
\textbf{Model} & \textbf{Demand Pattern} & \textbf{Horizon 14 Days} & \textbf{Horizon 30 Days} \\
\midrule
Croston & Erratic & 3.228 & 4.866 \\
& Intermittent & 1.999 & 2.975 \\
& Lumpy & 2.677 & 3.977 \\
& Smooth & 2.641 & 3.964 \\
& \textbf{Overall} & 2.568 & 3.837 \\
\midrule
LightGBM & Erratic & 3.265 & 5.051 \\
& Intermittent & 1.949 & 2.974 \\
& Lumpy & 2.793 & 4.416 \\
& Smooth & 2.346 & 3.531 \\
& \textbf{Overall} & 2.528 & 3.905 \\
\midrule
PatchTST & Erratic & 2.864 & 4.479 \\
& Intermittent & 1.877 & 2.833 \\
& Lumpy & 2.399 & 3.700 \\
& Smooth & 2.372 & 3.570 \\
& \textbf{Overall} & 2.323 & 3.555 \\
\midrule
Rule-based Router & Erratic & 2.723 & 4.270 \\
& Intermittent & 1.811 & 2.717 \\
& Lumpy & 2.172 & 3.357 \\
& Smooth & 2.369 & 3.579 \\
& \textbf{Overall} & 2.209 & 3.382 \\
\midrule
LightGBM Router & Erratic & 2.674 & 3.922 \\
& Intermittent & 1.790 & 2.839 \\
& Lumpy & 2.135 & 3.313 \\
& Smooth & 2.303 & 3.315 \\
& \textbf{Overall} & 2.168 & 3.282 \\
\midrule
InceptionTime Router & Erratic & \textbf{2.467} & \textbf{3.791} \\
& Intermittent & \textbf{1.748} & \textbf{2.633} \\
& Lumpy & \textbf{2.100} & \textbf{3.270} \\
& Smooth & \textbf{2.053} & \textbf{3.085} \\
& \textbf{Overall} & \textbf{2.051} & \textbf{3.134} \\
\bottomrule
\end{tabular}%
}
\label{tab:forecasting_performance}
\end{table}

\subsection{Classification Performance of Model Routers}
The ability of the LightGBM and InceptionTime routers to correctly identify the optimal forecasting model for each series was assessed using accuracy, macro-average F1, and weighted-average F1 scores, with results presented in Table~\ref{tab:classification_performance}.

The InceptionTime router demonstrated superior classification capabilities, particularly for the 14-day forecast horizon, achieving an accuracy of 0.43 and a weighted-average F1 score of 0.40. For the 30-day horizon, its accuracy remained at 0.43, though the macro-average F1 score was 0.28. The LightGBM router yielded accuracies of 0.34 (14-day) and 0.37 (30-day), with corresponding weighted-average F1 scores of 0.35 and 0.37.

While these absolute classification metrics might appear moderate, they are significantly better than random selection (which would be approximately 0.14, given seven models in the bank). More importantly, these results indicate that the routers are effectively learning to discern patterns that guide the selection towards better-performing models. The macro-average F1 scores, being generally lower than accuracy and weighted-average F1, suggest that predicting the optimal model for rarer or more difficult-to-distinguish series characteristics remains challenging.

Crucially, the forecasting performance improvements detailed in Section~\ref{sec:forecasting_performance}
demonstrate that even this level of classification accuracy is sufficient to yield substantial gains over any single-model approach. This underscores a key finding: the model router does not need to achieve perfect classification to be highly effective. By correctly guiding the model choice a significant portion of the time, or by selecting a 'good enough' model from the bank, the routers consistently enhance overall forecasting accuracy. This aligns with findings in meta-learning for time series forecasting, where imperfect but informed model selection has been shown to provide practical benefits \cite{talagala2018meta, montero2020fforma}.

\begin{table}[htbp]
\centering
\caption{Classification Performance of Model Routers}
\resizebox{\linewidth}{!}{%
\begin{tabular}{llccc}
\toprule
\textbf{Router Model} & \textbf{Forecast Horizon} & \textbf{Accuracy} & \textbf{Macro-avg F1} & \textbf{Weighted-avg F1} \\
\midrule
LightGBM Router & 14 days & 0.34 & 0.30 & 0.35 \\
 & 30 days & 0.37 & 0.33 & 0.37 \\
\midrule
InceptionTime Router & 14 days & 0.43 & 0.33 & 0.40 \\
 & 30 days & 0.43 & 0.28 & 0.37 \\
\bottomrule
\end{tabular}%
}
\label{tab:classification_performance}
\end{table}

\subsection{Computational Efficiency}
The inference runtimes for forecasting the 14-day horizon across all 5,000 holdout series are presented in Table~\ref{tab:runtime}. These results highlight significant differences in computational overhead between the various approaches.

Among the single-model benchmarks, Croston's method was the fastest (323.19 seconds), while the more complex LightGBM (13,124.86 seconds) and PatchTST (16,997.25 seconds) models incurred substantially higher computational costs.

Our model routers exhibited varied efficiencies. The \textbf{InceptionTime router} stands out, requiring only 3,637.81 seconds. This is notably faster than directly applying PatchTST (a reduction of approximately 78.6\%) or LightGBM (a reduction of approximately 72.3\%) to all series. This efficiency, coupled with its superior forecasting accuracy (as discussed in Section~\ref{sec:forecasting_performance}), positions the InceptionTime router as offering an excellent balance between predictive power and computational feasibility.

The Rule-based router also demonstrated reasonable efficiency (5,639.36 seconds), being faster than the standalone ML/DL benchmarks. In contrast, the LightGBM router (14,104.94 seconds) was comparable in runtime to using LightGBM as a direct forecaster, reflecting the overhead of feature computation and the LightGBM inference step for routing.

These findings underscore a key practical advantage: employing a sophisticated router like InceptionTime can lead to state-of-the-art forecasting accuracy without necessitating the deployment of the most computationally intensive models across every single time series. This efficient allocation of modeling resources is particularly valuable in large-scale supply chain forecasting environments.

\begin{table}[htbp]
\centering
\caption{Model Inference Time for 14-Day Forecast}
\resizebox{\linewidth}{!}{%
\begin{tabular}{lc}
\toprule
\textbf{Model} & \textbf{Computation Time (seconds)} \\
\midrule
Croston & 323.19 \\
LightGBM & 13,124.86 \\
PatchTST & 16,997.25 \\
Rule-based Router & 5,639.36 \\
LightGBM Router & 14,104.94 \\
InceptionTime Router & 3,637.81 \\
\bottomrule
\end{tabular}%
}
\label{tab:runtime}
\end{table}

\section{Discussion}
Our study provides strong evidence for the efficacy of model routing systems in tackling the challenges of sparse and intermittent demand forecasting. The primary insight is that dynamically selecting forecasting models based on time series characteristics yields significantly better accuracy than relying on any single model, even a state-of-the-art one like PatchTST. Specifically, our InceptionTime router reduced the overall NWRMSLE by approximately 12\% compared to the best single-model benchmark for both 14-day and 30-day horizon. This underscores the value of tailoring forecasting approaches to the diverse patterns inherent in large-scale retail datasets.

Interestingly, we found that even with moderate classification performance (accuracies around 0.34-0.43), the routers achieved substantial forecasting improvements. This suggests that the routing task benefits more from identifying a 'good' or 'better-than-average' model than from achieving perfect classification, which is a promising result for practical implementation. The success of the InceptionTime router, which leverages raw time series data, aligns with the growing consensus on the power of deep learning to capture complex temporal dynamics without extensive manual feature engineering \cite{dl_review_2025}. Our work extends these findings by demonstrating the successful application of a DL architecture (InceptionTime), originally noted for classification \cite{Ismail_Fawaz_2020}, to the \textbf{meta-task} of model selection, outperforming both feature-based ML (LightGBM router) and heuristic (Rule-based) approaches in both accuracy and computational efficiency. This comparative analysis across rule-based, ML, and DL routing methods addresses a notable gap in the existing literature, which often focuses on a single type of meta-learning strategy \cite{talagala2018meta, montero2020fforma}.

However, we acknowledge several limitations in our current study. Firstly, while our dataset is substantial, the training set size (approx. 20,000 series) might still constrain the full potential of deep learning routers; larger datasets could yield further improvements. Secondly, the 'best model' classification task is inherently challenging, especially given the potential for multiple models to perform similarly well on certain series, making the 'ground truth' somewhat ambiguous. Thirdly, the deep learning models within our \textbf{model bank} were trained with limited iterations (100) due to computational constraints, potentially underestimating their true forecasting capability and affecting the labels used for router training. Finally, our evaluation is based on a single, albeit large and complex, dataset (Favorita); further validation on datasets with different characteristics is warranted.

The problems or shortcomings we found in our study open up many interesting new questions and areas that researchers can explore in the future. Expanding the model bank to include a wider array of more diverse or fully-tuned forecasting techniques, particularly newer hybrid or DL models, would be a valuable next step. Developing more sophisticated routing architectures, potentially incorporating attention mechanisms or graph-based methods (if a relationship between series can be established), could further enhance selection accuracy. Exploring cost-sensitive learning for the routers, where misclassifying to a particularly poor model incurs a higher penalty, could improve robustness. Furthermore, implementing explainable AI (XAI) techniques could provide valuable insights into why a router selects a specific model, increasing trust and understanding. Additionally, since the LightGBM and InceptionTime routers output class probabilities, exploring their use to create weighted average ensemble forecasts—multiplying each model's forecast by its predicted probability—offers another promising direction beyond single-model selection. Finally, extending the framework to an online learning setting, where routers can adapt to evolving demand patterns over time, presents a significant opportunity for real-world supply chain applications.

\section{Conclusion}
This paper tackled sparse demand forecasting by systematically evaluating three model routing strategies: rule-based, LightGBM, and InceptionTime. On the large-scale Favorita dataset, all routers significantly outperformed single-model benchmarks. The InceptionTime router, using raw time series, notably achieved the best balance of forecasting accuracy and computational efficiency. Our findings demonstrate that intelligent model selection provides substantial forecasting improvements even with imperfect classification, highlighting the practical value of meta-learning. Future work will explore enhanced router architectures, expanded model banks, and online learning to further advance model routing systems for dynamic supply chains.

\bibliographystyle{ACM-Reference-Format}
\bibliography{references}


\begin{thebibliography}{18}


\ifx \showCODEN    \undefined \def \showCODEN     #1{\unskip}     \fi
\ifx \showDOI      \undefined \def \showDOI       #1{#1}\fi
\ifx \showISBNx    \undefined \def \showISBNx     #1{\unskip}     \fi
\ifx \showISBNxiii \undefined \def \showISBNxiii  #1{\unskip}     \fi
\ifx \showISSN     \undefined \def \showISSN      #1{\unskip}     \fi
\ifx \showLCCN     \undefined \def \showLCCN      #1{\unskip}     \fi
\ifx \shownote     \undefined \def \shownote      #1{#1}          \fi
\ifx \showarticletitle \undefined \def \showarticletitle #1{#1}   \fi
\ifx \showURL      \undefined \def \showURL       {\relax}        \fi
\providecommand\bibfield[2]{#2}
\providecommand\bibinfo[2]{#2}
\providecommand\natexlab[1]{#1}
\providecommand\showeprint[2][]{arXiv:#2}

\bibitem[Kag({[n.\,d.]})]%
        {Kaggle}
 \bibinfo{year}{[n.\,d.]}\natexlab{}.
\newblock \bibinfo{howpublished}{\url{https://www.kaggle.com/c/favorita-grocery-sales-forecasting}}.
\newblock


\bibitem[Chowdhury et~al\mbox{.}(2025)]%
        {dl_review_2025}
\bibfield{author}{\bibinfo{person}{Abdur Chowdhury}, \bibinfo{person}{Rajesh Paul}, {and} \bibinfo{person}{Farhana~Zaman Rozony}.} \bibinfo{year}{2025}\natexlab{}.
\newblock \showarticletitle{A SYSTEMATIC REVIEW OF DEMAND FORECASTING MODELS FOR RETAIL E-COMMERCE ENHANCING ACCURACY IN INVENTORY AND DELIVERY PLANNING}.
\newblock \bibinfo{journal}{\emph{International Journal of Scientific Interdisciplinary Research}}  \bibinfo{volume}{06} (\bibinfo{date}{03} \bibinfo{year}{2025}), \bibinfo{pages}{01--27}.
\newblock
\urldef\tempurl%
\url{https://doi.org/10.63125/mbbfw637}
\showDOI{\tempurl}


\bibitem[Croston(1972)]%
        {croston1972forecasting}
\bibfield{author}{\bibinfo{person}{J.D. Croston}.} \bibinfo{year}{1972}\natexlab{}.
\newblock \showarticletitle{Forecasting and stock control for intermittent demands}.
\newblock \bibinfo{journal}{\emph{Operational Research Quarterly}} \bibinfo{volume}{23}, \bibinfo{number}{3} (\bibinfo{year}{1972}), \bibinfo{pages}{289--303}.
\newblock


\bibitem[Eglite and Birzniece(2022)]%
        {dl_review_2022}
\bibfield{author}{\bibinfo{person}{Linda Eglite} {and} \bibinfo{person}{Ilze Birzniece}.} \bibinfo{year}{2022}\natexlab{}.
\newblock \showarticletitle{Retail Sales Forecasting Using Deep Learning: Systematic Literature Review}.
\newblock \bibinfo{journal}{\emph{Complex Systems Informatics and Modeling Quarterly}}  \bibinfo{volume}{0} (\bibinfo{date}{04} \bibinfo{year}{2022}), \bibinfo{pages}{53--62}.
\newblock
\urldef\tempurl%
\url{https://doi.org/10.7250/csimq.2022-30.03}
\showDOI{\tempurl}


\bibitem[Elman(1990)]%
        {RNN}
\bibfield{author}{\bibinfo{person}{Jeffrey~L. Elman}.} \bibinfo{year}{1990}\natexlab{}.
\newblock \showarticletitle{Finding structure in time}.
\newblock \bibinfo{journal}{\emph{Cognitive Science}} \bibinfo{volume}{14}, \bibinfo{number}{2} (\bibinfo{year}{1990}), \bibinfo{pages}{179--211}.
\newblock
\showISSN{0364-0213}
\urldef\tempurl%
\url{https://doi.org/10.1016/0364-0213(90)90002-E}
\showDOI{\tempurl}


\bibitem[Hyndman and Athanasopoulos(2018)]%
        {hyndman2018forecasting}
\bibfield{author}{\bibinfo{person}{{Robin John} Hyndman} {and} \bibinfo{person}{George Athanasopoulos}.} \bibinfo{year}{2018}\natexlab{}.
\newblock \bibinfo{booktitle}{\emph{Forecasting: Principles and Practice} (\bibinfo{edition}{2nd} ed.)}.
\newblock \bibinfo{publisher}{OTexts}, \bibinfo{address}{Australia}.
\newblock


\bibitem[Hyndman et~al\mbox{.}(2002)]%
        {HYNDMAN2002439}
\bibfield{author}{\bibinfo{person}{Rob~J Hyndman}, \bibinfo{person}{Anne~B Koehler}, \bibinfo{person}{Ralph~D Snyder}, {and} \bibinfo{person}{Simone Grose}.} \bibinfo{year}{2002}\natexlab{}.
\newblock \showarticletitle{A state space framework for automatic forecasting using exponential smoothing methods}.
\newblock \bibinfo{journal}{\emph{International Journal of Forecasting}} \bibinfo{volume}{18}, \bibinfo{number}{3} (\bibinfo{year}{2002}), \bibinfo{pages}{439--454}.
\newblock
\showISSN{0169-2070}
\urldef\tempurl%
\url{https://doi.org/10.1016/S0169-2070(01)00110-8}
\showDOI{\tempurl}


\bibitem[Ismail~Fawaz et~al\mbox{.}(2020)]%
        {Ismail_Fawaz_2020}
\bibfield{author}{\bibinfo{person}{Hassan Ismail~Fawaz}, \bibinfo{person}{Benjamin Lucas}, \bibinfo{person}{Germain Forestier}, \bibinfo{person}{Charlotte Pelletier}, \bibinfo{person}{Daniel~F. Schmidt}, \bibinfo{person}{Jonathan Weber}, \bibinfo{person}{Geoffrey~I. Webb}, \bibinfo{person}{Lhassane Idoumghar}, \bibinfo{person}{Pierre-Alain Muller}, {and} \bibinfo{person}{François Petitjean}.} \bibinfo{year}{2020}\natexlab{}.
\newblock \showarticletitle{InceptionTime: Finding AlexNet for time series classification}.
\newblock \bibinfo{journal}{\emph{Data Mining and Knowledge Discovery}} \bibinfo{volume}{34}, \bibinfo{number}{6} (\bibinfo{date}{Sept.} \bibinfo{year}{2020}), \bibinfo{pages}{1936–1962}.
\newblock
\showISSN{1573-756X}
\urldef\tempurl%
\url{https://doi.org/10.1007/s10618-020-00710-y}
\showDOI{\tempurl}


\bibitem[Ke et~al\mbox{.}(2017)]%
        {ke2017lightgbm}
\bibfield{author}{\bibinfo{person}{Guolin Ke}, \bibinfo{person}{Qi Meng}, \bibinfo{person}{Thomas Finley}, \bibinfo{person}{Taifeng Wang}, \bibinfo{person}{Wei Chen}, \bibinfo{person}{Weidong Ma}, \bibinfo{person}{Qiwei Ye}, {and} \bibinfo{person}{Tie-Yan Liu}.} \bibinfo{year}{2017}\natexlab{}.
\newblock \showarticletitle{LightGBM: a highly efficient gradient boosting decision tree}. In \bibinfo{booktitle}{\emph{Proceedings of the 31st International Conference on Neural Information Processing Systems}} (Long Beach, California, USA) \emph{(\bibinfo{series}{NIPS'17})}. \bibinfo{publisher}{Curran Associates Inc.}, \bibinfo{address}{Red Hook, NY, USA}, \bibinfo{pages}{3149–3157}.
\newblock
\showISBNx{9781510860964}


\bibitem[Makridakis and Hibon(2000)]%
        {m3_competition}
\bibfield{author}{\bibinfo{person}{Spyros Makridakis} {and} \bibinfo{person}{Michele Hibon}.} \bibinfo{year}{2000}\natexlab{}.
\newblock \showarticletitle{The M3-Competition: Results, Conclusions and Implications}.
\newblock \bibinfo{journal}{\emph{International Journal of Forecasting}}  \bibinfo{volume}{16} (\bibinfo{date}{10} \bibinfo{year}{2000}), \bibinfo{pages}{451--476}.
\newblock
\urldef\tempurl%
\url{https://doi.org/10.1016/S0169-2070(00)00057-1}
\showDOI{\tempurl}


\bibitem[Mohammadi(2021)]%
        {croston2021}
\bibfield{author}{\bibinfo{person}{Hamid Mohammadi}.} \bibinfo{year}{2021}\natexlab{}.
\newblock \bibinfo{title}{croston: Intermittent Demand Forecasting Methods in Python}.
\newblock \bibinfo{howpublished}{\url{https://pypi.org/project/croston/}}.
\newblock
\newblock
\shownote{Version 0.1.2.4}.


\bibitem[Montero-Manso et~al\mbox{.}(2020)]%
        {montero2020fforma}
\bibfield{author}{\bibinfo{person}{Pablo Montero-Manso}, \bibinfo{person}{George Athanasopoulos}, \bibinfo{person}{Rob~J. Hyndman}, {and} \bibinfo{person}{Thiyanga~S. Talagala}.} \bibinfo{year}{2020}\natexlab{}.
\newblock \showarticletitle{FFORMA: Feature-based forecast model averaging}.
\newblock \bibinfo{journal}{\emph{International Journal of Forecasting}} \bibinfo{volume}{36}, \bibinfo{number}{1} (\bibinfo{year}{2020}), \bibinfo{pages}{86--92}.
\newblock
\showISSN{0169-2070}
\urldef\tempurl%
\url{https://doi.org/10.1016/j.ijforecast.2019.02.011}
\showDOI{\tempurl}
\newblock
\shownote{M4 Competition}.


\bibitem[Nie et~al\mbox{.}(2023)]%
        {nie2023patchtst}
\bibfield{author}{\bibinfo{person}{Yuqi Nie}, \bibinfo{person}{Nam~H. Nguyen}, \bibinfo{person}{Phanwadee Sinthong}, {and} \bibinfo{person}{Jayant Kalagnanam}.} \bibinfo{year}{2023}\natexlab{}.
\newblock \bibinfo{title}{A Time Series is Worth 64 Words: Long-term Forecasting with Transformers}.
\newblock
\newblock
\showeprint[arxiv]{2211.14730}~[cs.LG]
\urldef\tempurl%
\url{https://arxiv.org/abs/2211.14730}
\showURL{%
\tempurl}


\bibitem[Olivares et~al\mbox{.}(2022)]%
        {olivares2022library_neuralforecast}
\bibfield{author}{\bibinfo{person}{Kin~G. Olivares}, \bibinfo{person}{Cristian Challú}, \bibinfo{person}{Federico Garza}, \bibinfo{person}{Max~Mergenthaler Canseco}, {and} \bibinfo{person}{Artur Dubrawski}.} \bibinfo{year}{2022}\natexlab{}.
\newblock \bibinfo{title}{{NeuralForecast}: User friendly state-of-the-art neural forecasting models.}
\newblock \bibinfo{howpublished}{{PyCon} Salt Lake City, Utah, US 2022}.
\newblock
\urldef\tempurl%
\url{https://github.com/Nixtla/neuralforecast}
\showURL{%
\tempurl}


\bibitem[Salinas et~al\mbox{.}(2019)]%
        {salinas2020deepar}
\bibfield{author}{\bibinfo{person}{David Salinas}, \bibinfo{person}{Valentin Flunkert}, {and} \bibinfo{person}{Jan Gasthaus}.} \bibinfo{year}{2019}\natexlab{}.
\newblock \bibinfo{title}{DeepAR: Probabilistic Forecasting with Autoregressive Recurrent Networks}.
\newblock
\newblock
\showeprint[arxiv]{1704.04110}~[cs.AI]
\urldef\tempurl%
\url{https://arxiv.org/abs/1704.04110}
\showURL{%
\tempurl}


\bibitem[Seabold and Perktold(2010)]%
        {seabold2010statsmodels}
\bibfield{author}{\bibinfo{person}{Skipper Seabold} {and} \bibinfo{person}{Josef Perktold}.} \bibinfo{year}{2010}\natexlab{}.
\newblock \showarticletitle{Statsmodels: Econometric and statistical modeling with python}. In \bibinfo{booktitle}{\emph{9th Python in Science Conference}}.
\newblock


\bibitem[Syntetos and Boylan(2005)]%
        {syntetos2005accuracy}
\bibfield{author}{\bibinfo{person}{Aris~A. Syntetos} {and} \bibinfo{person}{John~E. Boylan}.} \bibinfo{year}{2005}\natexlab{}.
\newblock \showarticletitle{The accuracy of intermittent demand estimates}.
\newblock \bibinfo{journal}{\emph{International Journal of Forecasting}} \bibinfo{volume}{21}, \bibinfo{number}{2} (\bibinfo{year}{2005}), \bibinfo{pages}{303--314}.
\newblock
\showISSN{0169-2070}
\urldef\tempurl%
\url{https://doi.org/10.1016/j.ijforecast.2004.10.001}
\showDOI{\tempurl}


\bibitem[Talagala et~al\mbox{.}(2023)]%
        {talagala2018meta}
\bibfield{author}{\bibinfo{person}{Thiyanga~S. Talagala}, \bibinfo{person}{Rob~J. Hyndman}, {and} \bibinfo{person}{George Athanasopoulos}.} \bibinfo{year}{2023}\natexlab{}.
\newblock \showarticletitle{Meta-learning how to forecast time series}.
\newblock \bibinfo{journal}{\emph{Journal of Forecasting}} \bibinfo{volume}{42}, \bibinfo{number}{6} (\bibinfo{year}{2023}), \bibinfo{pages}{1476--1501}.
\newblock
\urldef\tempurl%
\url{https://doi.org/10.1002/for.2963}
\showDOI{\tempurl}
\showeprint{https://onlinelibrary.wiley.com/doi/pdf/10.1002/for.2963}


\end{thebibliography}


\end{document}